\documentclass[runningheads]{llncs}

\usepackage{graphicx}
\usepackage{amsmath,amssymb}
\usepackage{color}
\usepackage{hyperref}
\begin{document}


\title{Toward Next-generation Medical Vision Backbones: Modeling Finer-grained Long-range Visual Dependency}
\titlerunning{Toward Next-generation Medical Vision Backbones}
\author{Mingyuan Meng}
\authorrunning{Mingyuan Meng}
\institute{School of Computer Science, The University of Sydney, Sydney, Australia \email{mmen2292@uni.sydney.edu.au}}

\maketitle
\begin{abstract}
Medical Image Computing (MIC) is a broad research topic covering both pixel-wise (e.g., segmentation, registration) and image-wise (e.g., classification, regression) vision tasks. Effective analysis demands models that capture both global long-range context and local subtle visual characteristics, necessitating fine-grained long-range visual dependency modeling. Compared to Convolutional Neural Networks (CNNs) that are limited by intrinsic locality, transformers excel at long-range modeling; however, due to the high computational loads of self-attention, transformers typically cannot process high-resolution features (e.g., full-scale image features before downsampling or patch embedding) and thus face difficulties in modeling fine-grained dependency among subtle medical image details. Concurrently, Multi-layer Perceptron (MLP)-based visual models are recognized as computation/memory-efficient alternatives in modeling long-range visual dependency but have yet to be widely investigated in the MIC community. This doctoral research advances deep learning-based MIC by investigating effective long-range visual dependency modeling. It first presents innovative use of transformers for both pixel- and image-wise medical vision tasks. The focus then shifts to MLPs, pioneeringly developing MLP-based visual models to capture fine-grained long-range visual dependency in medical images. Extensive experiments confirm the critical role of long-range dependency modeling in MIC and reveal a key finding: MLPs provide feasibility in modeling finer-grained long-range dependency among higher-resolution medical features containing enriched anatomical/pathological details. This finding establishes MLPs as a superior paradigm over transformers/CNNs, consistently enhancing performance across various medical vision tasks and paving the way for next-generation medical vision backbones.
\end{abstract}

\section{Research Problem and Motivation}
Modern medical image computing increasingly prioritizes network backbones capable of modeling long-range visual dependency, i.e., the ability to capture relationships between distant anatomical regions or features within or across images. This capability is essential as clinically significant visual patterns often span large areas or involve spatially dispersed structures~\cite{li2023transforming}. Effective modeling enhances global context awareness, improving identification of subtle abnormalities within complex anatomy and boosting robustness to anatomical variability. 

Convolutional Neural Networks (CNNs) initially dominated medical image computing due to their hierarchical feature extraction via translation-invariant convolutions. Their inherent strengths, such as local connectivity, weight sharing, and spatial hierarchy learning, are suitable for imaging data. However, their intrinsic locality fundamentally limits long-range dependency capture due to constrained receptive fields and absent global connectivity~\cite{li2021localvit}.

\begin{figure}[!t]
  \centering
  \includegraphics[width=0.75\textwidth]{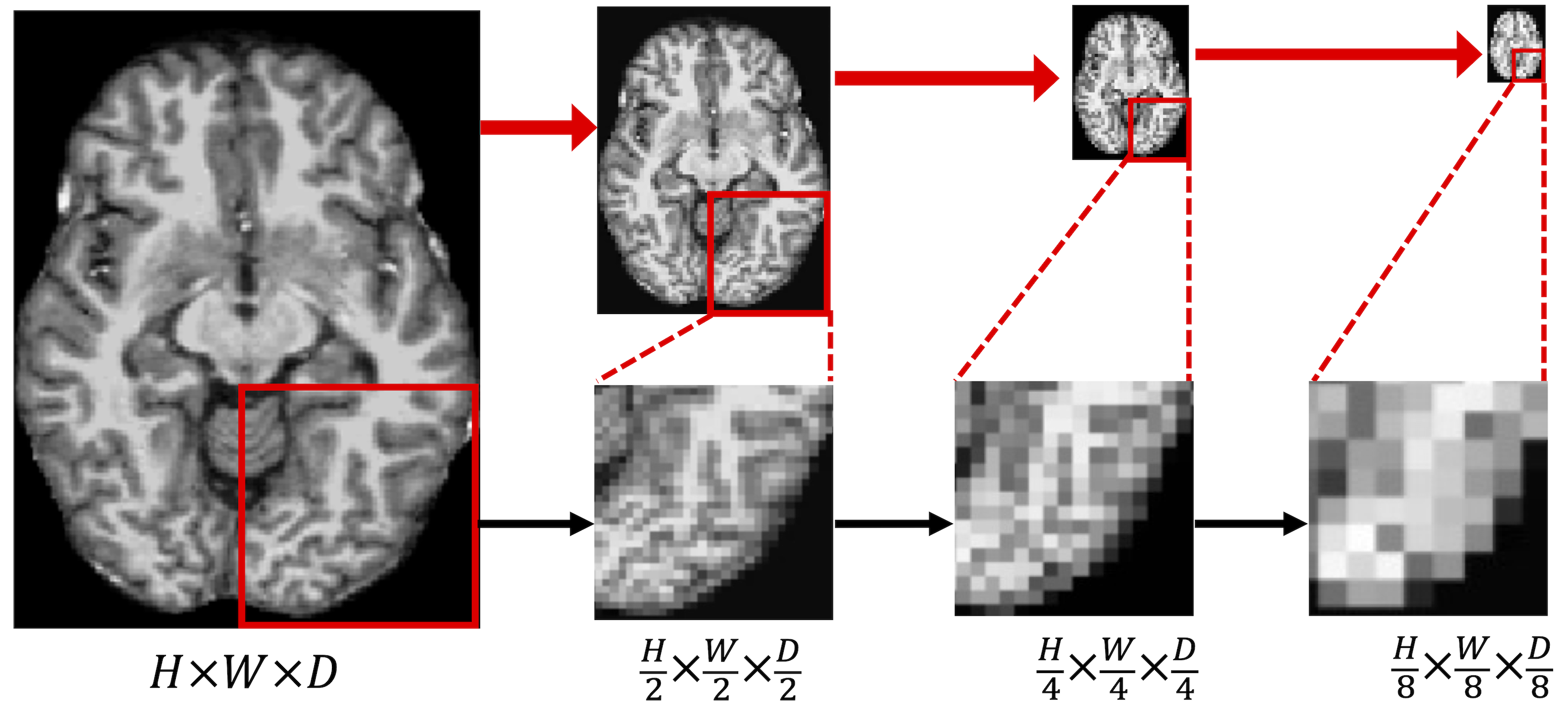}
  \caption{Illustration of a medical image after downsampling. A 2D slice of a 3D brain MRI scan shows white matter and grey matter (cortex) over the brain surface; the white and grey matter are blurred after downsampling and cannot be differentiated.}
  \label{fig:1}
\end{figure}

Transformers have recently emerged as a powerful alternative that leverages self-attention mechanisms from natural language processing~\cite{vaswani2017attention}. By enabling interactions between all input tokens (e.g., image patches), they effectively model global relationships, driving rapid adoption in medical imaging. However, the prohibitive computational costs of self-attention operations, especially for 3D medical imaging, remain key limitations~\cite{shamshad2023transformers}, incurring difficulties in processing high-resolution image features to capture fine-grained visual dependency~\cite{meng2023full}. While efficient variants (e.g., Swin Transformers~\cite{liu2021swin}) mitigate computational loads, they still fail to leverage the tissue-level textural information that is only available at high resolutions. Unfortunately, tissue-level textural details are indispensable for accurate medical image comprehension. Unlike natural images, medical images depict standardized anatomical regions across patients, resulting in high structural similarity with subtle anatomical/pathological characteristics discernible only in high resolutions containing tissue-level image textural details. As exemplified in Fig.~\ref{fig:1}, downsampled brain MRI lacks critical anatomical distinctions, e.g., gray and white matter boundaries. 

Under this context, Multi-layer Perceptrons (MLPs) serve as a promising frontier. MLP-based visual models can efficiently capture long-range visual dependency without costly self-attention operations\cite{liu2022we,tolstikhin2021mlp}. Their efficiency enables modeling of fine-grained long-range dependency among high-resolution features with critical subtle anatomical/pathological details. MLP-based models have been explored and achieved promising performance in natural image tasks, while their potential for medical image computing has not been fully recognized, as existing models lack the consideration of inductive bias (also known as learning bias) crucial for medical image computing.

This doctoral research~\cite{meng2025modeling} aims to uncover the fundamental mechanisms by which long-range visual dependency enhances deep learning-based medical image computing, thereby advancing the field through more effective long-range dependency modeling. This research pioneers novel transformer- and MLP-based network architectures to effectively model long-range visual dependency and rigorously evaluates them across diverse medical image computing tasks. 

\section{Background}
Medical image computing has become a pivotal component of modern healthcare, fueled by the critical role of medical imaging in disease diagnosis and prognosis. Medical imaging technologies, such as Magnetic Resonance Imaging (MRI), Computed Tomography (CT), Positron Emission Tomography (PET), and X-rays, generate detailed visualizations of internal anatomy and physiological processes, which are indispensable for detecting abnormalities, evaluating disease severity, and designing personalized treatment strategies. The primary objective of medical image computing is to derive diagnostic and prognostic insights from medical images, empowering clinicians to make precise decisions and develop personalized treatment plans~\cite{gu2023multi,gu2022prediction}. This field encompasses diverse vision tasks, which can be broadly categorized as pixel-wise and image-wise vision tasks: Pixel-wise (also known as dense prediction) tasks require pixel-level predictions for fine-grained analysis, including anatomical structure segmentation~\cite{ye2024enabling} or image registration~\cite{meng2022enhancing}, while image-wise tasks derive holistic interpretations from medical images for disease classification~\cite{li2025enhancing} or outcome prediction~\cite{meng2021multi}. The clinical importance and widespread nature of medical imaging have motivated intense research and clinical efforts on computational medical image analysis.

In recent years, deep learning has established itself as a transformative force in medical image computing, demonstrating remarkable success across both pixel-wise and image-wise vision tasks~\cite{suganyadevi2022review}. Unlike traditional machine learning approaches that demand handcrafted feature engineering and domain-specific expertise, deep learning automatically extracts high-level pattern representations directly from medical images through deep neural networks. This capability eliminates human bias inherent in handcrafted feature engineering while uncovering clinically relevant semantic features potentially missed by manually-defined feature extraction~\cite{hosny2019handcrafted}. The advancement of deep learning has been propelled by increased computational power, expanded datasets, and novel network architectures. Despite these developments, significant architectural challenges persist for medical applications. Medical images exhibit substantial variability across patients and pathologies, necessitating architectures with global perception capabilities to contextualize anatomical structures and suppress irrelevant local variations. Simultaneously, the critical diagnostic importance of subtle textural details demands precise localized perception to capture fine-grained anatomical and pathological features. Therefore, the pursuit of deep learning-based medical image computing has driven significant architectural evolution, especially in visual backbones capable of long-range visual dependency modeling. 

Transformers, with global self-attention mechanisms, have demonstrated substantial success in various medical vision tasks. For medical image segmentation, deep learning models such as TransUNet~\cite{chen2021transunet} integrated vision transformers into UNet bottlenecks to enhance contextual awareness. For deformable registration, TransMorph~\cite{chen2022transmorph} leveraged Swin transformers~\cite{liu2021swin} to capture spatial relationships across images. Despite these advances, transformers face persistent challenges: their computational complexity remains prohibitive for high-resolution 3D images, and even optimized variants (e.g., Swin transformer) struggle to preserve fine-grained anatomical details in full-resolution feature maps. In image-wise tasks such as survival prediction, transformer-based models, e.g., MCTA~\cite{chen2021multimodal}, capture prognostic patterns but often inadequately address multi-modal fusion, relying on simplistic early fusion rather than cross-modal interaction.

Concurrently, MLP-based visual models have emerged as efficient alternatives for long-range dependency modeling. Early models (e.g., MLP-Mixer~\cite{tolstikhin2021mlp}) excelled in classification but lacked multi-scale support for pixel-wise dense prediction. Later hierarchical MLP-based models (e.g., Hire-MLP~\cite{guo2022hire}) enabled pixel-wise tasks, while MAXIM~\cite{tu2022maxim} achieved state-of-the-art natural image processing performance by handling high-resolution features. For medical images, the adoption of MLPs remains limited. UNeXt~\cite{valanarasu2022unext} accelerated medical image segmentation with token-shifted MLPs, while a few studies explored other tasks such as registration~\cite{wang2022unsupervised} and reconstruction~\cite{li20243dpx}. Existing approaches adopted generic MLP models without domain-specific adaptations, lacking the consideration of inductive bias (i.e., initial assumptions about the data to be analyzed/generalized) that is crucial for target tasks. Further, existing approaches initiate MLP processing after feature downsampling (e.g., 4×4 patch embedding), discarding tissue-level textures essential for precise medical image comprehension. These limitations highlight the unmet need for medical-optimized MLP frameworks capturing fine-grained long-range dependency at high resolutions. 

\section{Scientific Approach}
As illustrated in Fig.~\ref{fig:2}, this doctoral research advances long-range visual dependency modeling for medical image computing through four studies:

\begin{figure}[!t]
  \centering
  \includegraphics[width=\textwidth]{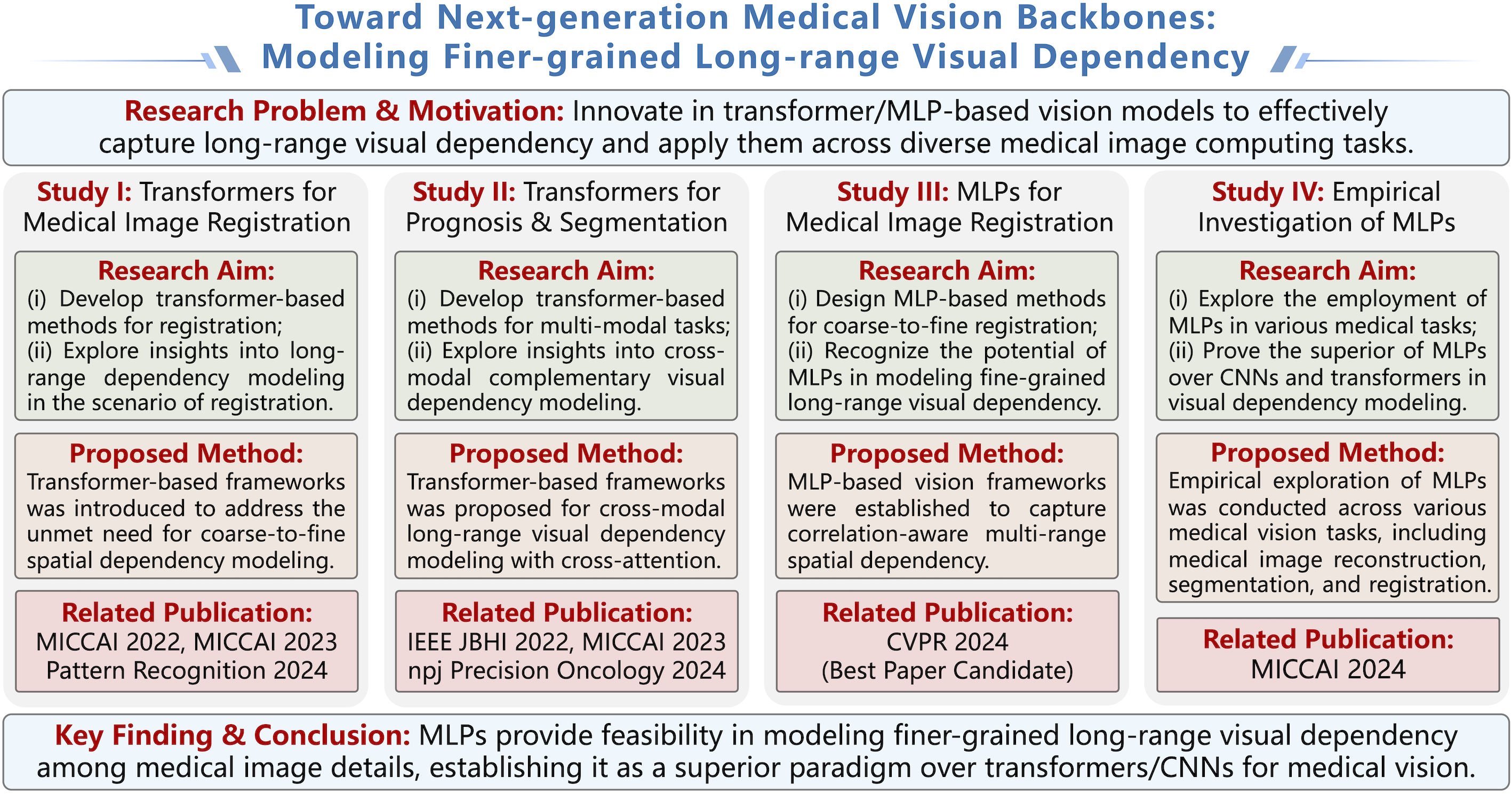}
  \caption{Overview of this doctoral research.}
  \label{fig:2}
\end{figure}

\textbf{Study I: Transformers for medical image registration} - This study aims at (i) developing a new transformer-based method for medical image registration and (ii) exploring new insights into how the modeling of long-range visual dependency works in the scenario of image registration. To attain these aims, a transformer-based medical image registration framework (named NICE-Trans) was proposed to address the unmet need for coarse-to-fine dependency modeling. Distinct from previous frameworks, NICE-Trans embeds transformers into a non-iterative coarse-to-fine registration framework to model long-range relevance between images. This framework enables joint modeling of affine and deformable transformations while capturing long-range spatial relevance across resolutions, establishing new benchmarks in registration tasks. The research related to this study has been published at \textit{MICCAI 2022}~\cite{meng2022non}, \textit{MICCAI 2023}~\cite{meng2023non}, and \textit{Pattern Recognition}~\cite{meng2025autofuse}.

\textbf{Study II: Transformers for survival prediction and segmentation} - This study aims at (i) developing a new transformer-based method for joint survival prediction and tumor segmentation from multi-modality PET-CT images and (ii) exploring new insights into how transformers benefit representation learning by modeling long-range complementary dependency between multi-modality medical images. To attain these aims, a merging-diverging hybrid transformer network (named XSurv) was proposed, where a Hybrid Parallel Cross-Attention (HPCA) block was introduced to effectively integrate complementary PET-CT information via cross-attention transformers. This model enables extracting modality-specific prognostic features while sharing contextual knowledge across modalities. The research related to this study has been published at \textit{IEEE JBHI}~\cite{meng2022deepmts}, \textit{MICCAI 2023}~\cite{meng2023merging}, and \textit{npj Precision Oncology}~\cite{meng2024adaptive}.

\textbf{Study III: MLPs for medical image registration} - This study aims at (i) developing the first MLP-based method for medical image registration and (ii) revealing the unnoticed potential of MLPs in modeling fine-grained long-range visual dependency for pixel-wise image registration tasks. To attain these aims, a correlation-aware coarse-to-fine MLP-based network (named CorrMLP) was proposed, which introduces the first MLP block that was designed to model correlation-aware multi-range visual dependency for medical image registration. This model unlocks the potential of MLPs for capturing pixel-wise spatial dependency. This study has been published at \textit{CVPR 2024}~\cite{meng2024correlation} and was nominated as the \textit{Best Paper Candidate} (Top 24/0.2\%).

\textbf{Study IV: Empirical investigation of MLPs in medical dense prediction} - This study aims at (i) exploring the advantages of MLPs in modeling fine-grained long-range visual dependency among high-resolution image features and (ii) validating their superiority over transformers in medical image computing. To attain these aims, a comprehensive empirical study was conducted to explore the employment of MLPs on various pixel-wise prediction tasks, including medical image reconstruction, registration, and segmentation. This empirical study also introduces a new feature extraction framework that produces hierarchical feature pyramids with fine-grained long-range dependency modeling and can be generalized to a wide range of medical dense prediction tasks. The research innovating in MLP-based models for medical image reconstruction has been published at \textit{MICCAI 2024}~\cite{li20243dpx}.

\section{Proposed Solution}
Study I introduces NICE-Trans, a non-iterative coarse-to-fine transformer network that unifies affine and deformable medical image registration within a single network. This study makes two technical advances: (i) It extended the coarse-to-fine registration paradigm to jointly model traditionally separated affine and deformable transformations in one network iteration, eliminating multi-stage processing. (ii) It innovatively embedded transformers to progressively capture long-range spatial relevance between images, marking the first integration of transformers into non-iterative coarse-to-fine registration. These dual advances establish NICE-Trans as a pioneering model to unify affine and deformable coarse-to-fine registration while capturing long-range spatial dependency. 

Study II introduces XSurv, an X-shaped merging-diverging hybrid transformer network for joint survival prediction and tumor segmentation. The architecture of XSurv features a merging encoder that fuses complementary anatomical (from CT) and metabolic (from PET) information, and a diverging decoder that extracts region-specific prognostic features from primary tumor and metastatic lymph node regions. The technical innovations comprise: (i) a specialized merging-diverging learning framework for joint survival prediction and tumor segmentation, enabling multi-modality exploitation and region-specific feature extraction applicable across survival tasks; (ii) a Hybrid Parallel Cross-Attention (HPCA) block that concurrently learns local intra-modality features through convolutional pathways and global inter-modality dependencies via cross-attention transformers; and (iii) a Region-specific Attention Gate (RAG) that filters lesion-relevant features through anatomical screening.

Study III introduces CorrMLP, the first MLP-based coarse-to-fine deformable registration framework. Its core innovation is the correlation-aware multi-window MLP (CMW-MLP) block, a purpose-built module that computes local feature correlations and captures multi-range spatial dependency through parallel MLP operations. This block was embedded into a novel correlation-aware architecture leveraging both image-level (inter-image feature relationships) and step-level (inter-scale contextual propagation) correlations. This dual-correlation mechanism provides enriched contextual guidance throughout the coarse-to-fine registration process. CorrMLP thus pioneers three technical contributions: (i) establishing MLPs as promising backbones for deformable registration; (ii) introducing the first registration-optimized MLP block with explicit correlation modeling; and (iii) devising a contextual registration framework where both image-level and step-level correlations actively guide registration.

Study IV uncovers the underexplored capability of MLPs in capturing fine-grained long-range dependency in high-resolution image features, a critical advantage for medical dense prediction. Through a comprehensive empirical investigation, a hierarchical MLP framework was introduced, which extracts multi-scale feature pyramids via hierarchical MLP blocks operating beginning from the full image resolution. Task-specific decoders then leverage these feature pyramids for various medical applications, including medical image reconstruction, deformable registration, and segmentation. Crucially, when evaluating various MLP blocks within this framework, a paradigm-shifting finding was observed: regardless of the specific MLP variants, employing MLPs at full resolution consistently enabled superior performance over CNN- and transformer-based methods across all evaluation tasks, even outperforming task-optimized specialist models.

\section{Results and Contribution}
The methodologies proposed in this research were extensively validated across multi-modal medical imaging datasets (including MRI, CT, and PET) and body regions (including brain, cardiac, and head and neck regions), leveraging public benchmarks ~\cite{eisenmann2022biomedical,baheti2021brain,meng2022radiomics,meng2022brain} to ensure reproducibility and clinical relevance.

In Study I, the proposed NICE-Trans achieved state-of-the-art performance in medical image registration by unifying affine and deformable coarse-to-fine registration within a single non-iterative network~\cite{meng2023non}. In quantitative evaluations, NICE-Trans attained the best Dice Similarity Coefficient (DSC) results across all evaluation datasets, outperforming existing transformer-based methods and CNN-based coarse-to-fine methods. Qualitatively, its alignments show superior anatomical consistency with fixed images. The architecture’s joint affine-deformable design proves computationally optimal: affine registration incurs negligible runtime overhead compared to standalone affine methods, while end-to-end training reduces GPU memory burdens versus multi-network pipelines. Ablation studies reveal that embedding transformers in the decoder, not the encoder, drives performance gains by modeling inter-image spatial relevance rather than intra-image representations. This insight counters prior transformer-based registration methods (e.g., TransMorph) and establishes a new principle: transformers in the decoder maximize registration efficacy by explicitly modeling long-range spatial correspondence to handle large deformations between images.

In Study II, the proposed XSurv achieved state-of-the-art performance in head and neck cancer, attaining the highest C-index for survival prediction on the HECKTOR 2022 challenge dataset while simultaneously outperforming multi-task models in both survival prediction and tumor segmentation~\cite{meng2023merging}. This dual superiority stems from its novel architecture: the merging encoder with Hybrid Parallel Cross-Attention (HPCA) blocks solves critical PET-CT fusion limitations by concurrently extracting intra-modality features and inter-modality relevance, outperforming typical early and late fusion strategies; while the diverging decoder with Region-specific Attention Gates (RAG) achieves precise localization of primary tumors and metastatic lymph nodes, evidenced by attention maps and overall best segmentation DSC among all comparison methods. Crucially, XSurv eliminates reliance on manual segmentation during prognosis, a significant advantage over traditional radiomics-based methods requiring anatomical priors. This integrated approach establishes a new paradigm for joint multi-modal survival modeling, where optimized feature interaction and region-specific decoding drive superior prognostic performance.

In Study III, the proposed CorrMLP established state-of-the-art deformable registration performance, exceeding existing CNN-based, transformer-based, and coarse-to-fine registration methods in both brain and cardiac image datasets~\cite{meng2024correlation}. It achieved significantly higher DSC while maintaining competitive transformation smoothness and real-time GPU processing (\(<\)1s per image pair), with qualitative results demonstrating exceptional anatomical alignment. Three innovations drive these performance gains: (i) The proposed correlation-aware coarse-to-fine framework leverages image-level feature relationships and step-level contextual propagation; (ii) Full-resolution MLP processing outperforms CNNs/transformers by capturing fine-grained dependency in high-resolution features; and (iii) The CMW-MLP block employs optimized multi-window operations (3×3×3, 5×5×5, 7×7×7) to jointly model subtle and large deformations, outperforming five existing MLP variants.

In Study IV, the conducted empirical study validated MLPs as superior vision backbones for medical dense prediction, demonstrating state-of-the-art performance across reconstruction, registration, and segmentation tasks~\cite{meng2023full}. For low-dose PET reconstruction, MLP-based MLP-Unet surpassed transformer-based methods by 3.2-8.7\% across metrics, with the largest gains in ultra-low dose conditions where sparse textures challenge conventional methods. For deformable registration, MLP-based MLPMorph outperformed transformer-based methods by 2.1-4.3\% DSC while maintaining real-time speed, resolving registration's critical reliance on high-resolution textural details. For segmentation, MLP-based MLP-Unet also achieved the highest DSC, excelling particularly at boundary delineation as shown in qualitative analysis. Crucially, ablation studies revealed two paradigm-shifting insights: (i) The advantage of MLPs stems from high-resolution long-range dependency modeling: replacing first-stage MLPs with convolutions degraded all tasks, while transformers failed at full resolution due to GPU constraints; (ii) Performance gains are architecture-agnostic: four distinct MLP blocks outperformed transformers when applied at high resolution, proving the inherent efficacy of MLPs over transformers for fine-grained dependency modeling. This study might redefine the backbone selection principle: under the same computational constraints, MLPs can unlock finer-grained long-range dependency modeling for pixel-wise medical analysis. 

\section{Open Challenges and Future Work}
It is hoped that this doctoral research can inspire new discussions on the employment of MLPs in medical image computing and motivate the community to recognize the potential of MLPs as a superior technical paradigm over CNNs and transformers for capturing finer-grained long-range visual dependency in medical images. However, despite the technical advances in this doctoral research, key barriers remain for clinical adoption: Firstly, future work could prioritize multi-site robustness validation. Large-scale clinical trials using heterogeneous multi-center data are essential to verify performance under real-world variability. Secondly, addressing data bias and fairness is critical. Future efforts could focus on curating inclusive datasets and developing fairness-aware adaptations of MLP networks to ensure equitable performance across populations. Finally, model interpretability remains a critical issue for clinical trustworthiness. Future work could pioneer task-specific interpretability methods, e.g., visualizing fine-grained dependency maps in MLP-based frameworks.

\section{Long Term Goals}
The long-term goals are to (i) motivate the community to rethink the critical roles of long-range vision dependency modeling in medical image computing and (ii) drive a paradigm shift in medical vision backbones by advancing MLPs as a superior alternative to transformers in long-range visual dependency modeling.

\bibliographystyle{splncs04}
\bibliography{main}

\end{document}